\documentclass[USenglish,twocolumn]{article}
\usepackage[utf8]{inputenc}
\usepackage[big,online]{dgruyter}
\usepackage[sort&compress,square,numbers]{natbib}
\usepackage{times}
\usepackage{subfig}
\usepackage{xcolor}
\usepackage[detect-all=true,separate-uncertainty=true, multi-part-units=single,range-phrase={-},product-units = single]{siunitx}
\DeclareSIUnit{\pixel}{px}
\DeclareSIUnit{\fps}{fps}

\usepackage{pgfplots}
\usepgfplotslibrary{groupplots}
\pgfplotsset{width=10cm,compat=1.9}

\usepackage{verbatim}
\usepackage{tikz}
\usetikzlibrary{arrows,positioning,shapes.geometric}

\usetikzlibrary{intersections}

\begin{document}

  \DOI{10.1515/}
  \openaccess
  \pagenumbering{gobble}

\title{Robust Tracking with Particle Filtering for Fluorescent Cardiac Imaging}
\runningtitle{Robust Tracking with Particle Filtering for Fluorescent Cardiac Imaging}

\author*[1]{S.~Guttikonda}
\author[1]{M.~Neidhardt}
\author[1]{J.~Sprenger}
\author[2]{J.~Petersen}
\author[2]{C.~Detter}
\author[1,5]{A.~Schlaefer}

\runningauthor{S.~Guttikonda et al.}

\affil[1]{\protect\raggedright 
  Institute of Medical Technology and Intelligent Systems, Hamburg University of Technology, Am Schwarzenberg-Campus 3, Hamburg, Germany, e-mail: suresh.guttikonda@tuhh.de}
  
\affil[2]{\protect\raggedright 
Department of Cardiovascular Surgery, University Heart and Vascular Center
Hamburg, Hamburg, Germany}

\affil[5]{\protect\raggedright 
 SustAInLivWork Center of Excellence, Kaunas, Lithuania}

\abstract{Intraoperative fluorescent cardiac imaging enables quality control following coronary bypass grafting surgery. We can estimate local quantitative indicators, such as cardiac perfusion, by tracking local feature points. However, heart motion and significant fluctuations in image characteristics caused by vessel structural enrichment limit traditional tracking methods. We propose a particle filtering tracker based on cyclic-consistency checks to robustly track particles sampled to follow target landmarks. Our method tracks $117$ targets simultaneously at \SI{25.4}{\fps}, allowing real-time estimates during interventions. It achieves a tracking error of (\SI{5.00(2.2)}{\pixel}) and outperforms other deep learning trackers (\SI{22.3(11)}{\pixel}) and conventional trackers (\SI{58.1(27.1)}{\pixel}).}



\maketitle

\section{Introduction}
Fluorescent cardiac imaging (FCI) offers a non-invasive technique for visualizing coronary vessels. FCI provides real-time, dynamic information about cardiac function to diagnose pathological conditions, such as coronary bypass stenosis~\cite{detter2018qualitative} or graft assessment~\cite{ohmes2017techniques}. Additionally, clinicians can estimate quantitative values based on fluorescence intensity; for example, to determine myocardial perfusion~\cite{detter2018qualitative}. To achieve this, robustly tracking local features on the deforming and moving heart surface presents a significant challenge and remains highly desirable.

Conventional markerless object tracking methods, such as Minimum Output Sum of Squared Error (MOSSE)~\cite{DBLP:conf/cvpr/BolmeBDL10} and Kernelized Correlation Filter (KCF)~\cite{DBLP:conf/eccv/HenriquesCMB12}, operate with computational efficiency and enable real-time tracking. However, these methods deliver limited tracking accuracy when objects experience partial or full occlusion, images contain artifacts (e.g., motion blur caused by rapid movements), or image features change (e.g., during contrast enrichment).

\begin{figure}
    \centering
    \includegraphics[width=\linewidth]{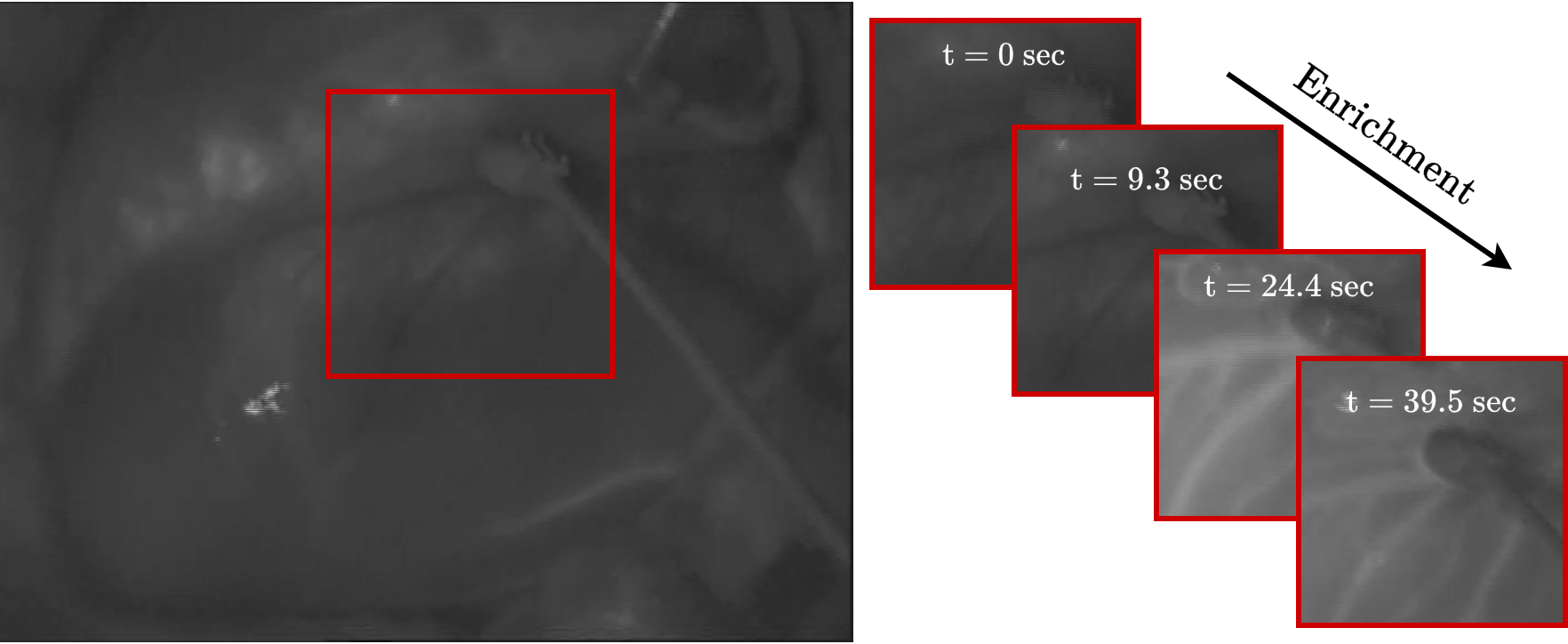}
    \caption{\textbf{Fluorescent Cardiac Imaging:} Video sequences illustrate different stages of the heart enrichment procedure. Large motions and varying image features challenge the effectiveness of conventional tracking approaches.}
    \label{img:heart_tracking_data}
\end{figure}

Recently, tracking any point (TAP) approaches~\cite{DBLP:conf/iccv/DoerschYVG0ACZ23, DBLP:journals/corr/abs-2410-11831} have gained attention due to its ability to estimate both dense and long-range pixel-level trajectories in video sequences, even with occlusions. These techniques fundamentally use a coarse-to-fine tracking design that iteratively predicts an initial "coarse" point trajectories using low-resolution features and then refines it using local, spatiotemporal information at higher resolution. Researchers train these methods either with unlabeled raw data combined with unsupervised or semi-supervised learning~\cite{DBLP:journals/corr/abs-2410-11831}, or with large-scale synthetic data through supervised learning~\cite{DBLP:conf/iccv/DoerschYVG0ACZ23}.

\begin{figure*}
    \centering
    \subfloat[Amid Enrichment] {
        \begin{tikzpicture}
            \node[anchor=south west,inner sep=0] (img) at (0, 0){\includegraphics[trim=0 0 0 0, clip, width=0.495\linewidth]{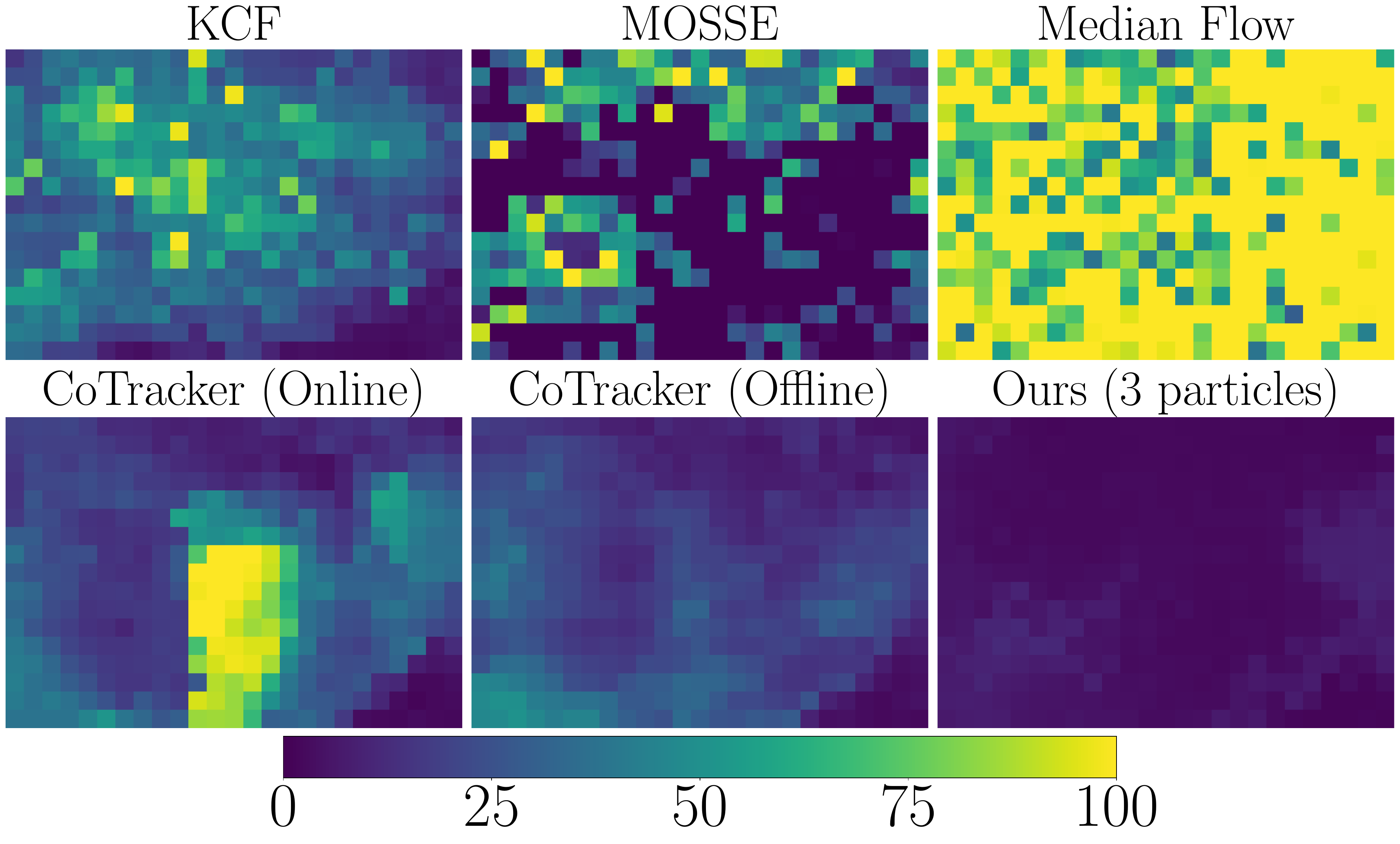}}; 
            \node[] at (0.75,0.4) {\scriptsize FBE [px]}; 
        \end{tikzpicture} 
        \label{img:tracking_map_amid_enrichment}
    }
    \subfloat[Post Enrichment] {
        \begin{tikzpicture}
            \node[anchor=south west,inner sep=0] (img) at (0, 0){\includegraphics[trim=0 0 0 0, clip, width=0.495\linewidth]{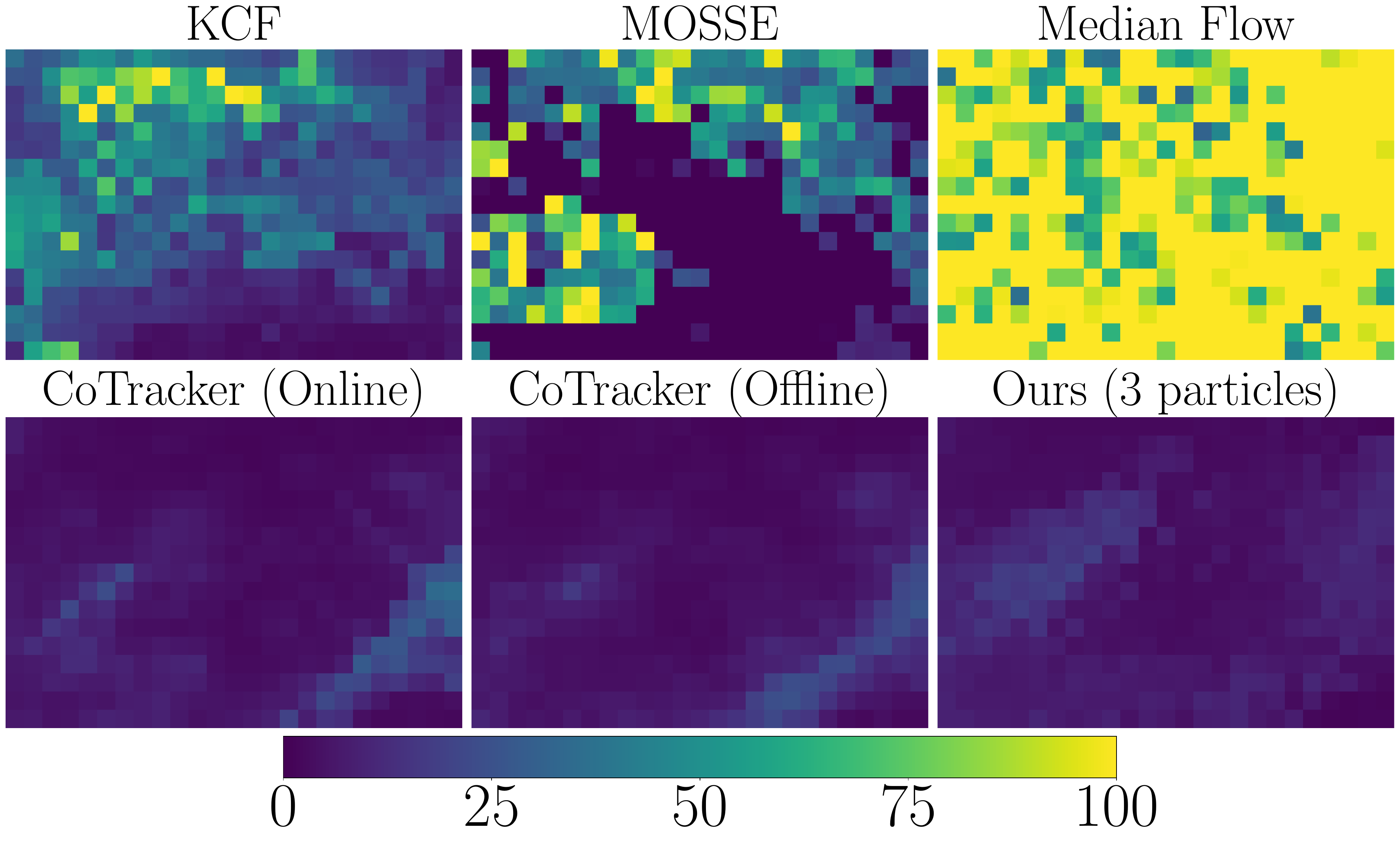}}; 
            \node[] at (0.75,0.4) {\scriptsize FBE [px]}; 
        \end{tikzpicture}
        \label{img:tracking_map_post_enrichment}
   }
    \caption{\textbf{Tracking Error Dissemination:} We present the distribution of the FBE for all evaluated trackers. Tracking occurs (a) during contrast agent injection - amid enrichment (AE), and (b) post enrichment (PE), on video sequences with a duration of \SI{15}{\second}, respectively.}
 \end{figure*}

In this work, we propose to retain the benefits of the TAP algorithm, trained on a large-scale dataset, while achieving real-time tracking performance without fine-tuning by refining its predictions using a particle filter–style filtering mechanism. Our study shows that the proposed method outperforms both conventional and deep learning trackers and exhibits greater robustness to large appearance and scale changes during both amid enrichment (AE) and post enrichment (PE) scenarios.

\vspace{-5mm}
\section{Method}

\subsection{Notations} \label{sec:formulation}
Given a video $\begin{Bmatrix}\boldsymbol{\mathcal{I}}_{t}\end{Bmatrix}^{T}_{t=1}$, which consists of a sequence of $\text{T}$ frames $\boldsymbol{\mathcal{I}}_{t} \in \mathbb{R}^{H \times W \times 3}$, and an anchor point $\boldsymbol{\mathcal{Q}} = (t^{q}, x^{q}, y^{q}) \in \mathbb{R}^{3}$, where $t^{q}$ denoting the query frame index, and $(x^{q}, y^{q})$ representing the corresponding spatial location of the anchor point, the point tracking problem aims to estimate the trajectory of this anchor point across the video. Specifically, the forward point track comprises predictions on subsequent frames of the image sequence and is formally denoted as $\boldsymbol{\mathcal{P}}^{f}_{t} = (x_{t}, y_{t}) \in \mathbb{R}^{2}$, $t \in [1, T]$, with $(x_{t}, y_{t}) = (x^{q}, y^{q})$.

Correspondingly, we define a backward point track consisting of predictions performed from the last to the first frame of the sequence. It is formally denoted as $\boldsymbol{\mathcal{P}}^{b}_{t} = (\hat{x}_{t}, \hat{y}_{t}) \in \mathbb{R}^{2}$, $t \in [1, T]$, with $(\hat{x}_{T}, \hat{y}_{T}) = (x_{T}, y_{T})$.

\subsection{Our Tracking Algorithm} \label{sec:our_approach}

We propose to track the center of a bounding box with size $w \times w$~pixels. First, we sample $M$ points, referred to as particles, $\begin{Bmatrix}\boldsymbol{X}^{m}_{0}\end{Bmatrix}^{M}_{m=1} \in \mathbb{R}^{2}$, from a Gaussian distribution $ \mathcal{N}\begin{pmatrix}\boldsymbol{X}_{0} = [{x}^{q}, {y}^{q}]^{\top}; {\sigma}_{0} = 5.0\end{pmatrix} $ centered at the bounding box center. Each particle receives an initial weight, defined by forward-backward tracking consistency~\cite{DBLP:conf/icpr/KalalMM10, DBLP:conf/cvpr/WangJE19}, such that $\begin{Bmatrix}{w}^{m}_{0}\end{Bmatrix}^{M}_{m=1} = \frac{1}{M}$, where ${\sum}^{M}_{m=1} w^{m}_{t} = 1$ and $t \in [0, T]$.

Next, we compute forward prediction tracks for the sampled particles across a sequence of $L$ frames. We recognize that particles experience varying tracking accuracy due to the highly deformable heart surface, which causes occlusions and changes in image features from contrast enrichment. To maintain robustness, we remove particles based on their corresponding weights. Following this, we estimate the Gaussian likelihood ($\mathcal{L}$) for each particle as,
\begin{multline}
\mathcal{L}(\boldsymbol{\mathcal{P}}^{b}_{t}, \boldsymbol{\Sigma} | \boldsymbol{\mathcal{P}}^{f}_{t}) =  \\ \frac{1}{2 \pi \sqrt{|\boldsymbol{\Sigma}|}} exp\begin{pmatrix} - \frac{1}{2} 
                        (\boldsymbol{\mathcal{P}}^{b}_{t} - \boldsymbol{\mathcal{P}}^{f}_{t})^{\top}\boldsymbol{\Sigma}^{-1}(\boldsymbol{\mathcal{P}}^{b}_{t} - \boldsymbol{\mathcal{P}}^{f}_{t})\end{pmatrix}
\label{eq:neg_loglikelihood_error}
\end{multline}
with the covariance matrix $\boldsymbol{\Sigma} = \begin{bmatrix} \sigma^2 & 0 \\ 0 & \sigma^2 \end{bmatrix}$, the standard deviation $\sigma = 3.0$, $d$ represent the data dimensions, and $|\boldsymbol{\Sigma}|$ denote the determinant of the covariance matrix.

Finally, we perform stochastic universal soft-resampling\footnote[1]{\url{https://github.com/stanford-iprl-lab/torchfilter}}~\cite{DBLP:books/daglib/0014221, DBLP:conf/corl/KarkusHL18} with a trade-off resampling parameter $\alpha = 0.5$. This process retains or replaces particles with lower weights by generating new particles sampled near those with higher weights at the end of each filtering window, thereby enhancing tracking robustness and accuracy.

\vspace{-5mm}
\subsection{Experimental Evaluation}
We evaluate different tracking approaches using $14$ videos recorded during pig heart fluorescent cardiac interventions. Following anesthesia, the pigs undergo coronary revascularization on their beating hearts and median sternotomy. During the procedure, we inject a contrast agent (indocyanine green) and expose the heart to near-infrared light at \SI{785}{\nano\meter}. The fluorescence signal is captured using an FSI device (LLS GmbH, Ulm, Germany) equipped with band-pass filters that block excitation light while selectively transmitting fluorescent light at \SI{830}{\nano\meter}. Each recording lasts approximately \SI{100}{\second}, with a temporal resolution of \SI{25}{\Hz} and a spatial resolution of \SI{512}{\pixel} $\times$ \SI{384}{\pixel}. Figure~\ref{img:heart_tracking_data} illustrates a complete enrichment cycle, including both amid and post enrichment sequences. We conduct all training and evaluations using a GPU (Titan RTX, Nvidia, California, USA).

To evaluate tracking performance, we perform cyclic-consistency checks between the forward and backward predicted point tracks~\cite{DBLP:conf/icpr/KalalMM10}. We define the forward-backward tracking error (FBE) for each anchor point as follows:

\begin{equation}
FBE =  \sum^{T}_{t=0} || \boldsymbol{\mathcal{P}}^{b}_{t} - \boldsymbol{\mathcal{P}}^{f}_{t} ||_2
\label{eq:forward_backward_error}
\end{equation}
where $||\cdot||_2$ is the Euclidean norm.

\begin{figure}
    \centering
    \begin{tikzpicture}
        \node[anchor=south west,inner sep=0] (img) at (0,0){\includegraphics[trim=0 16cm 7cm 0, clip, width=\linewidth]{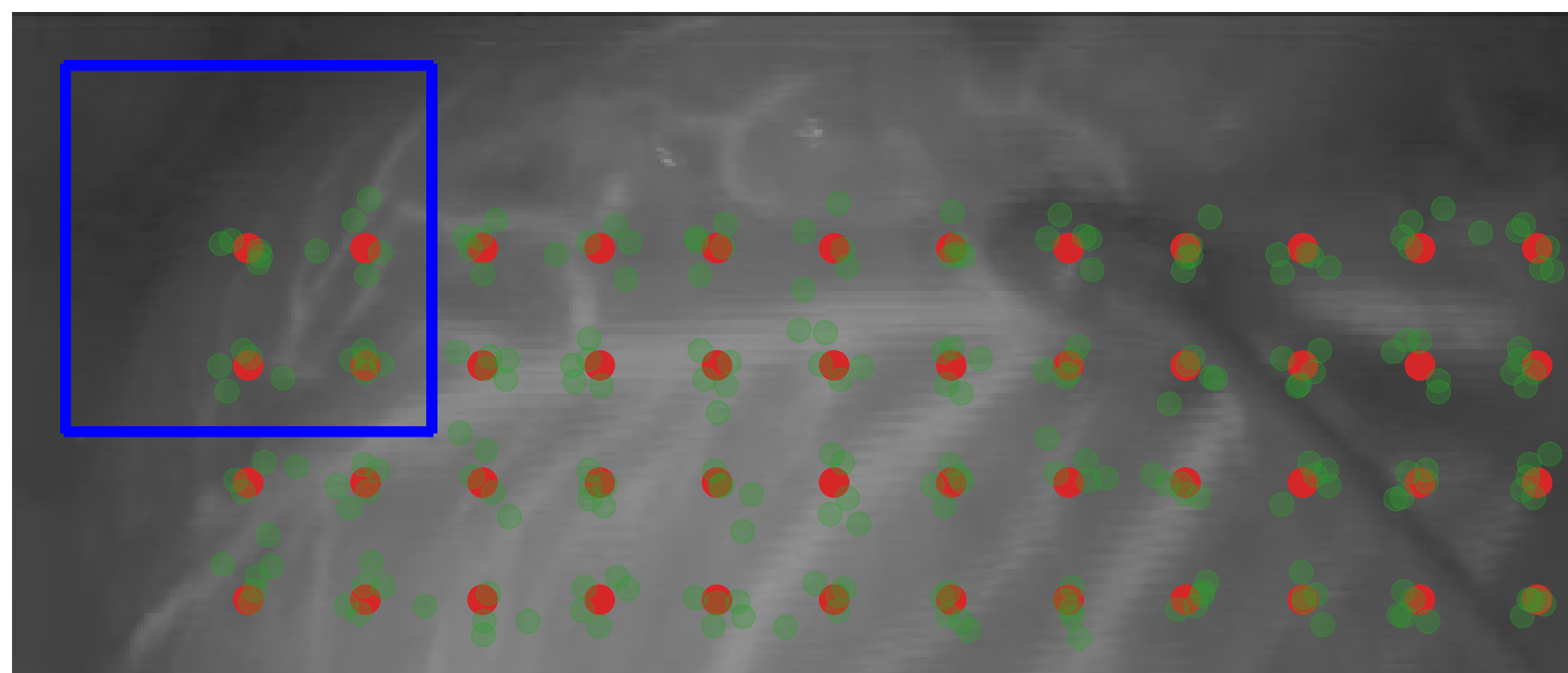}};
   
        \node[] at (1.3cm,3.0cm) {\textcolor{white}{\scriptsize\SI{100}{\pixel} $\times$ \SI{100}{\pixel}}}; 
        \draw [|<->|,>=stealth, thick, white] (3.2,3.7-1.0) -- (3.8,3.7-1.0) node [midway,above] {\textcolor{white}{\scriptsize $\SI{32}{\pixel}$}};

    \end{tikzpicture}
    \caption{\textbf{Evaluation:} We evaluate tracking performance using bounding boxes (blue) arranged in a grid structure. The red dots represent anchor points positioned at the center of each bounding box. For our proposed tracker, we sample particles (green dots) located within the bounding boxes.}
    \label{img:dense_grid_anchors_bboxes_particles}
\end{figure}

As shown in Figure~\ref{img:dense_grid_anchors_bboxes_particles}, we estimate tracking performance on a grid of anchor points spaced \SI{32}{\pixel} apart along both axes. We place the sample anchor points at the center of bounding boxes with size w=\SI{100}{\pixel}, as smaller sizes result in immediate tracking failures. For comparison, we select conventional bounding box–based trackers\footnote[2]{\url{https://docs.opencv.org/3.4.20/d0/d0a/classcv_1_1Tracker.html}} including MOSSE~\cite{DBLP:conf/cvpr/BolmeBDL10}, KCF~\cite{DBLP:conf/eccv/HenriquesCMB12}, and Median Flow~\cite{DBLP:conf/icpr/KalalMM10}. For our proposed tracker, we sample a collection of particles from a Gaussian distribution around each anchor point. We ensure that all sampled particles lie within the bounding box to provide a fair comparison with other methods. We evaluate our tracker using $M\in{[3,5,25]}$ sampled particles and set the filtering window length to $L = 16$ frames for all experiments.

\vspace{-5mm}
\section{Results} \label{sec:our_results}

First, as shown in Figures~\ref{img:tracking_map_amid_enrichment} and~\ref{img:tracking_map_post_enrichment}, we assess the distribution of the FBE, defined in Equation~\ref{eq:forward_backward_error}, on the surface of the heart during amid-enrichment (AE) and post-enrichment (PE) sequences, respectively. The results reveal that traditional trackers (top row) exhibit a non-homogeneous FBE distribution; for example, MOSSE tracks only discrete features on the heart surface with low FBE. In PE tracking, deep learning–based trackers provide more consistent estimates, whereas they only partially succeed during AE tracking (see CoTracker online). Our approach delivers robust estimates that remain consistent regardless of tracking location, achieving a mean FBE of \SI{4.9(4.8)}{\pixel} and \SI{4.4(3.6)}{\pixel} for AE and PE sequences, respectively.


\begin{figure*}
    \centering
    \includegraphics[width=\linewidth]{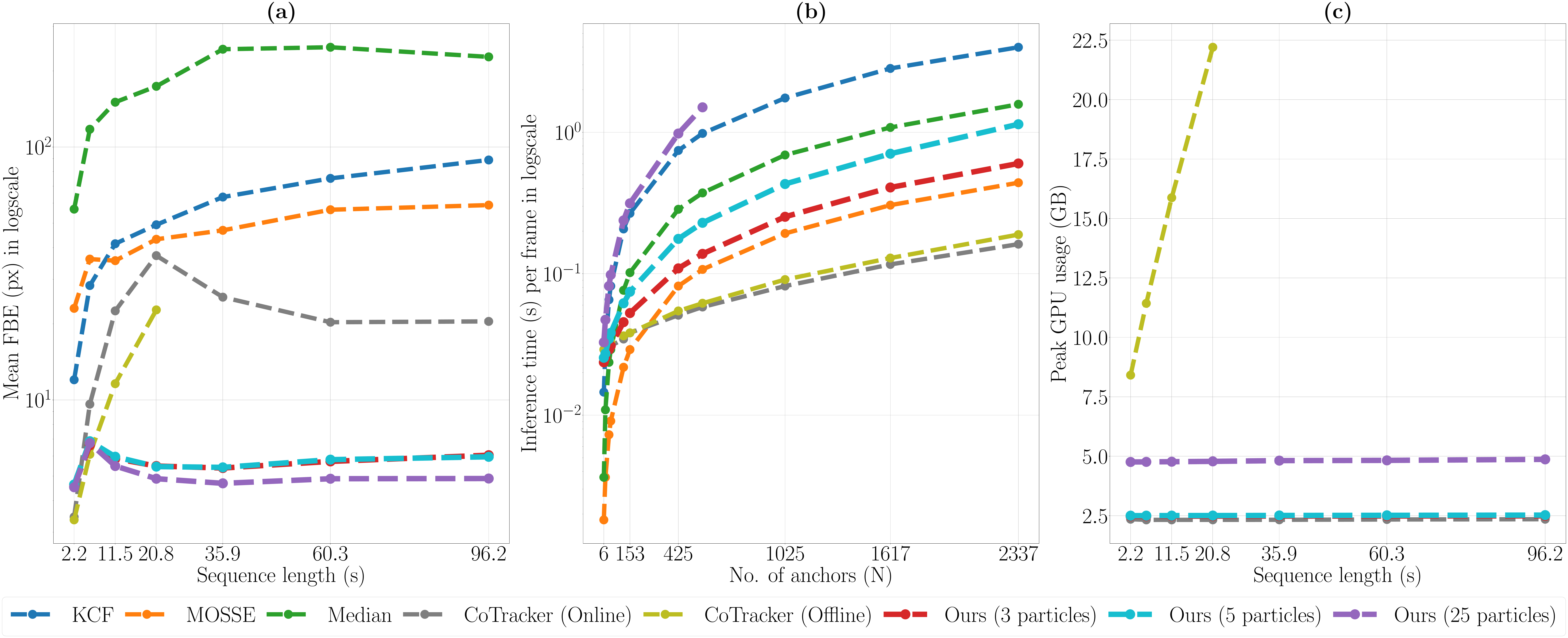}
    \caption{\textbf{Tracking Performance of All Trackers:} (a) Tracking accuracy vs sequence length, (b) Inference time vs the number of anchor points, and (c) Peak GPU memory usage vs sequence length.}

    \label{img:Performance_Trackers}
\end{figure*}

\begin{table}[!ht]
\footnotesize
\caption{\textbf{Tracking Performance:} We report the mean FBE and frames per second (fps) for tracking $117$ anchor points across $14$ videos during contrast agent injection, amid enrichment (AE), and post enrichment (PE), each with a tracking duration of \SI{15}{\second}. We refer to the total tracking duration of \SI{90}{\second}, which includes both AE and PE sequences, as "Total".}
\renewcommand\arraystretch{1.3}
\renewcommand\tabcolsep{0pt}
  \begin{tabular*}{\linewidth}{@{\extracolsep{\fill}}lccccc}
    \hline  
    \text{}   &  & \multicolumn{3}{c}{FBE}   \\  \cmidrule{3-5}
    Tracker               & fps           & AE                      & PE                      &  Total                  \\ \midrule
    KCF                   &  5.2          &\SI{24.3(13.7)}{}        & \SI{26.0(17.3)}{}       &\SI{58.1(27.1)}{}        \\
    MOSSE                 & \textbf{50.8} &\SI{38.3(17.9)}{}        &\SI{43.8(21.0)}{}        &\SI{70.8(17.8)}{}        \\
    Median Flow           & 13.9          &\SI{84.7(49.9)}{}        &\SI{83.5(47.9)}{}        &\SI{159.9(59.2)}{}       \\
    CoTracker (Online)    & 42.5          &\SI{25.8(19.9)} {}       &\SI{6.4(3.4)}{}          &\SI{22.3(11.0)}{}        \\
    CoTracker (Offline)   & 31.3          &\SI{15.4(10.6)}{}        &\SI{5.7(3.4)}{}          & --                      \\ \midrule
    Ours (3 particles)    & \textbf{25.4} &\SI{4.9(4.8)}  {}        &\SI{4.4( 3.6)}{}         &\SI{5.0(2.2)}{}          \\
    Ours (5 particles)    & 17.9          &\SI{4.8(4.6)} {}         &\SI{4.3( 3.4)}{}         &\SI{4.8(2.1)}{}          \\
    Ours (25 particles)   & 4.2           &\textbf{\SI{4.5(4.7)}{}} &\textbf{\SI{3.9(3.3)}{}} &\textbf{\SI{4.3(2.2)}{}} \\
    \hline
  \end{tabular*}\par\medskip
\label{tab:dense_grid_tracking_rmse}
\end{table}

Additionally, we calculate the average FBE for image sequences of varying durations, as shown in Figure~\ref{img:Performance_Trackers}a. Compared to other trackers, our approach achieves a notably lower mean FBE for tracking sequences longer than \SI{10}{\second}. Figure~\ref{img:Performance_Trackers}a demonstrates that while our tracking error remains consistently below \SI{7}{\pixel} during AE, the mean FBE significantly increases for all other trackers. During PE, the error varies only marginally across all methods. A similar trend appears in Table~\ref{tab:dense_grid_tracking_rmse}, where conventional trackers show overall higher tracking errors compared to our method. Notably, unlike our approach, CoTracker’s tracking error increases only slightly during PE sequences but nearly triples during AE sequences. Furthermore, increasing the number of particles reduces the FBE in our approach. With just three particles, our method outperforms all other trackers, achieving an overall FBE of \SI{5.0(2.2)}{\pixel}.

Compared to CPU-based traditional trackers, as shown in Figure~\ref{img:Performance_Trackers}b, our approach exhibits a longer inference time. However, the inference time per anchor point decreases as we reduce the number of particles. Achieving a frame rate of up to \SI{25.4}{\fps} with three particles—matching the fluorescent cardiac image stream—we can track $117$ anchor points in parallel while maintaining a lower FBE compared to other trackers. Figure~\ref{img:Performance_Trackers}c presents the GPU peak consumption for various tracking sequence lengths by deep learning–based trackers. While our method remains stable over longer sequences, it incurs slightly higher computational costs than CoTracker (online) due to the use of $M$ particles per anchor point, although it remains more efficient than CoTracker (offline).

\vspace{-5mm}
\section{Discussion and Conclusion}

Real-time capture of rapid movements and changes on the heart’s surface is essential for estimating quantitative metrics and assessing heart functions. While conventional trackers offer computational efficiency, they exhibit poor tracking accuracy during fluorescent cardiac imaging (FCI). Advanced deep learning–based trackers, such as CoTracker, demonstrate improved tracking performance with robustness to occlusions; however, they still struggle to overcome domain gaps and face trade-offs between accuracy and GPU computational constraints over longer sequences. In this work, we employ a particle filter mechanism to refine CoTracker’s predictions, enabling robust real-time estimates even when image features change over time, such as during contrast agent injection.

Our experiments indicate that, when used out-of-the-box, CoTracker (offline) outperforms conventional trackers but is limited to video sequences of up to \SI{21}{\second}. In contrast, CoTracker (online) can handle longer sequences but achieves lower overall performance. By applying our proposed approach, we achieve a fourfold improvement in tracking performance on \SI{90.0}{\second}-long video sequences while simultaneously tracking 117 anchor points at a temporal sampling rate of \SI{25.0}{\fps}.



\noindent
\textsf{\textbf{Author Statement}}\\
This research was co-funded by the MARLOC project (DFG, grant SCHL 1844-10-1) and by the European Union under Horizon Europe programme grant agreement No. 101059903; and by the European Union funds for the period 2021-2027. Conflict of interest: none. Informed consent: obtained. Ethical approval: not applicable.
 

\vspace{-7mm}
\bibliographystyle{elsevier3}
\bibliography{Proceedings_template_LaTeX}

\newpage

\end{document}